\documentclass[10pt,twocolumn,letterpaper]{article}

\usepackage{iccv}
\usepackage{times}
\usepackage{epsfig}
\usepackage{graphicx}
\usepackage{amsmath}
\usepackage{amssymb}

\usepackage{xcolor}
\usepackage{booktabs}
\usepackage{tabularx}
\usepackage{colortbl}
\usepackage{textcomp}
\usepackage{verbatim}
\usepackage{multirow}
\usepackage{makecell}
\usepackage{subcaption}
\usepackage{siunitx}
\usepackage{authblk}

\makeatletter  
\@namedef{ver@everyshi.sty}{}
\renewcommand\AB@affilsepx{\hskip 0.15in \protect\Affilfont}

\makeatother
\usepackage{pgf}

\usepackage[pagebackref=true,breaklinks=true,letterpaper=true,colorlinks,bookmarks=false]{hyperref}

\usepackage[capitalize]{cleveref}

 \iccvfinalcopy 

\def\iccvPaperID{****} 

\ificcvfinal\fi

\definecolor{LightGray}{gray}{0.95}

\newcommand{\textapprox}{\raisebox{0.5ex}{\texttildelow}}

\usepackage{color}
\usepackage[absolute]{textpos}
\definecolor{gray92}{gray}{0.92}

\begin{document}



\begin{textblock*}{\textwidth}(18mm,12mm)
    \textblockcolour{gray92}
    \vspace{2mm}
    \tiny
    \centering
    Lorenzo Brigato, Bj{\"o}rn Barz, Luca Iocchi and Joachim Denzler.\\
    ``Tune It or Don't Use It: Benchmarking Data-Efficient Image Classification.''\\
    \textit{Proceedings of the IEEE/CVF International Conference on Computer Vision (ICCV) Workshops 2021.}\\
    \copyright\ 2021 IEEE. Personal use of this material is permitted.
    Permission from IEEE must be obtained for all other uses, in any current or future media, including reprinting/republishing this material for advertising or promotional purposes, creating new collective works, for resale or redistribution to servers or lists, or reuse of any copyrighted component of this work in other works.
    The final publication will be available at
    \href{https://ieeexplore.ieee.org/}{ieeexplore.ieee.org}.\\
    \vspace{2mm}
\end{textblock*}

\title{Tune It or Don't Use It: \\ Benchmarking Data-Efficient Image Classification}

\author[1]{Lorenzo Brigato\thanks{{\tt\footnotesize brigato@diag.uniroma1.it}}}
\author[2]{Bj{\"o}rn Barz\thanks{{\tt\footnotesize bjoern.barz@uni-jena.de}}}
\author[1]{Luca Iocchi}
\author[2]{Joachim Denzler}
\affil[1]{Sapienza University of Rome}
\affil[2]{Friedrich Schiller University Jena}

\maketitle
\ificcvfinal\thispagestyle{empty}\fi

\begin{abstract}
Data-efficient image classification using deep neural networks in settings, where only small amounts of labeled data are available, has been an active research area in the recent past.
However, an objective comparison between published methods is difficult, since existing works use different datasets for evaluation and often compare against untuned baselines with default hyper-parameters.
We design a benchmark for data-efficient image classification consisting of six diverse datasets spanning various domains (e.g., natural images, medical imagery, satellite data) and data types (RGB, grayscale, multispectral).
Using this benchmark, we re-evaluate the standard cross-entropy baseline and eight methods for data-efficient deep learning published between 2017 and 2021 at renowned venues.
For a fair and realistic comparison, we carefully tune the hyper-parameters of all methods on each dataset.
Surprisingly, we find that tuning learning rate, weight decay, and batch size on a separate validation split results in a highly competitive baseline, which outperforms all but one specialized method and performs competitively to the remaining one.
\end{abstract}

\section{Introduction}
\label{sec:intro}

Many recent advances in computer vision and machine learning in general have been achieved by large-scale pre-training on massive datasets \cite{dosovitskiy2021vit,cui2018large,radford2019gpt}.
As the amount of data grows, the importance of methodological advances vanishes.
With the number of training samples approaching infinity, a simple k-nearest neighbor classifier provides optimal performance \cite{torralba2008tinyimages}.
The true hallmark of intelligence is, therefore, the ability of learning generalizable concepts from limited amounts of data.

The research area of \emph{deep learning from small data} or \emph{data-efficient deep learning} has been receiving increasing interest in the past couple of years \cite{oyallon2017scaling, barz2020deep, brigato2021close, bruintjes2021vipriors}.
However, an objective comparison of proposed methods is difficult due to the lack of a common benchmark.
Even if two works use the same dataset for evaluation, their random sub-samples of this dataset for simulating a small-data scenario will be different and not directly comparable.

Fortunately, there recently have been activities to establish common benchmarks and organize challenges to foster direct competition between proposed methods \cite{bruintjes2021vipriors}.
Still, they are often limited to a single dataset, \eg, ImageNet \cite{russakovsky2015imagenet}, which comprises a different type of data than usually encountered in a small-data scenario.

\begin{figure}[t]
    \resizebox{\linewidth}{!}{\input{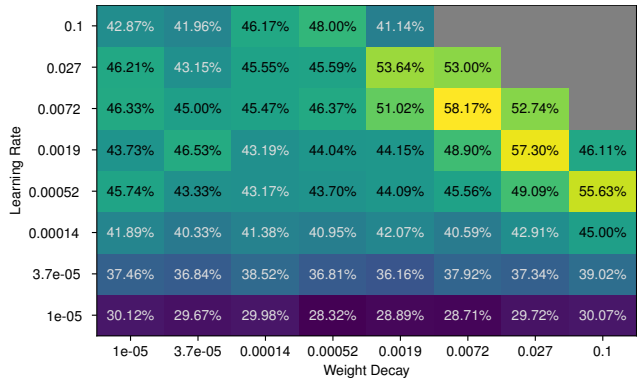}}
    \caption{Classification accuracy obtained with standard cross-entropy on ciFAIR-10 with 1\% of the training data for different combinations of learning rate and weight decay. Gray configurations led to divergence.}
    \label{fig:gridsearch}
\end{figure}

Moreover, most existing works compare their proposed method against insufficiently tuned baselines \cite{barz2020deep} or baselines trained with default hyper-parameters \cite{oyallon2017scaling,ulicny2019harmonic,kayhan2020translation,sun2020visual,kobayashi2021t}, which makes it easy to outperform them.
However, careful tuning of hyper-parameters, as one would do in practice, is crucial and can have a considerable impact on the final performance \cite{bischl2021hpo}, as illustrated in \cref{fig:gridsearch}.
Here, we evaluated the performance of several combinations of learning rate and weight decay for a standard cross-entropy classifier with a Wide ResNet architecture \cite{zagoruyko2016wrn} trained on as few as 1\% of the CIFAR-10 training data \cite{krizhevsky2009learning} and evaluated on the ciFAIR-10 test set \cite{barz2020cifair} (see \cref{sec:setup} for details on the training procedure).
Typical default hyper-parameters such as a learning rate of 0.1 and weight decay of \num{1e-4} as used by \cite{bruintjes2021vipriors} would achieve \textapprox 46\% accuracy in this scenario, which is entire 12 percentage points below the optimal performance of \textapprox 58\%.
Even works that do perform hyper-parameter tuning often only optimize the learning rate and keep the weight decay fixed to some default from $[\num{1e-5}, \num{1e-4}]$.
Such a procedure results in similarly suboptimal performance on this small training dataset, which apparently requires much stronger regularization.
We can furthermore observe that the best performing hyper-parameter combinations are close to an area of the search space that results in divergence of the training procedure.
This makes hyper-parameter optimization a particularly delicate endeavor.

In this work, we establish a direct, objective, and informative comparison by re-evaluating the state of the art in data-efficient image classification.
To this end, we introduce a comprehensive benchmark consisting of six datasets from a variety of domains: natural images of everyday objects, fine-grained classification, medical imagery, satellite images, and handwritten documents.
Two datasets consist of non-RGB data, where the common large-scale pre-training and fine-tuning procedure is not straightforward, emphasizing the need for methods that can learn from limited amounts of data from scratch.
To facilitate evaluating novel methods, we share the dataset splits of our benchmark under \url{https://github.com/cvjena/deic}.

Using this benchmark, we re-evaluate eight selected state-of-the-art methods for data-efficient image classification.
The hyper-parameters of all methods are carefully optimized for each dataset individually on a validation split, while the final performance is evaluated on a separate test split.
Surprisingly and somewhat disillusioning, we find that thorough hyper-parameter optimization results in a strong baseline, which outperforms seven of the eight specialized methods published in the recent literature.

In the following, we first introduce the datasets constituting our benchmark in \cref{sec:datasets}.
Then, we briefly describe the methods selected for the comparison in \cref{sec:methods}.
Our experimental setup and training procedure are detailed in \cref{sec:setup} and the results are presented in \cref{sec:results}.
\Cref{sec:conclusions} summarizes the conclusions from our study.

\section{Datasets}
\label{sec:datasets}

Most works on deep learning from small datasets use custom sub-sampled versions of popular standard image classification benchmarks such as ImageNet \cite{russakovsky2015imagenet} or CIFAR \cite{krizhevsky2009learning}.
This limited variety bears the risk of overfitting research progress to individual datasets and the domain covered by them, in this case, photographs of natural scenes and everyday objects.
In particular, this is not the domain typically dealt with in a small-data scenario, where specialized data that is difficult to obtain or annotate is in the focus.
Additionally, very recent work showed that high performance on ImageNet does not necessarily correlate to high performance on other vision datasets \cite{tuggener2021enough}.

Therefore, we compile a diverse benchmark consisting of six datasets from a variety of domains and with different data types and numbers of classes.
We sub-sampled all datasets to fit the small-data regime, with the exception of CUB \cite{wah2011cub}, which was already small enough.
By default, we aimed for 50 training images per class.
This full \emph{trainval} split is only used for the final training and furthermore split into a training (\textapprox 60\%) and a validation set (\textapprox 40\%) for hyper-parameter optimization.
For testing the final models trained on the trainval split, we used official standard test datasets where they existed.
Only for two datasets, namely EuroSAT \cite{helber2019eurosat} and ISIC 2018 \cite{codella2019isic}, we had to create own test splits.
A summary of the dataset statistics is given in \cref{tab:datasets}.
In the following, we briefly describe each individual dataset used for our benchmark.
Example images from all datasets are shown in \cref{fig:dataset-examples}.

\begin{table*}[t]
    \centering
    \renewcommand{\arraystretch}{1.3}
    \setlength{\tabcolsep}{10pt}
    \begin{tabularx}{\linewidth}{X r r r r l l}
        \toprule
         Dataset                                                & Classes & Imgs/Class & \#Trainval &  \#Test  & Problem Domain & Data Type \\
         \midrule
         ImageNet-1k \cite{russakovsky2015imagenet}             &   1,000 &         50 &    50,000 &  50,000 & Natural Images & RGB \\
         \rowcolor{LightGray}
         ciFAIR-10 \cite{krizhevsky2009learning,barz2020cifair} &      10 &         50 &       500 &  10,000 & Natural Images & RGB (32x32) \\
         CUB \cite{wah2011cub}                                  &     200 &         30 &     5,994 &   5,794 & Fine-Grained   & RGB \\
         \rowcolor{LightGray}
         EuroSAT \cite{helber2019eurosat}                       &      10 &         50 &       500 &  19,500 & Remote Sensing & Multispectral \\
         ISIC 2018 \cite{codella2019isic}                       &       7 &         80 &       560 &   1,944 & Medical        & RGB \\
         \rowcolor{LightGray}
         CLaMM \cite{stutzmann2016clamm}                        &      12 &         50 &       600 &   2,000 & Handwriting    & Grayscale \\
         \bottomrule
    \end{tabularx}
    \caption{Datasets constituting our benchmark. Except for CUB, we use sub-samples to simulate a small-data scenario.}
    \label{tab:datasets}
\end{table*}

\begin{figure*}[t]
    \setlength{\tabcolsep}{0pt}
    \renewcommand{\arraystretch}{0}
    \begin{subfigure}[b]{.334\linewidth}%
        \offinterlineskip%
        \resizebox{\linewidth}{!}{%
            \includegraphics[height=4cm]{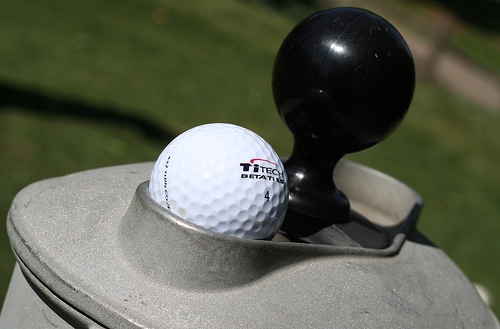}%
            \includegraphics[height=4cm]{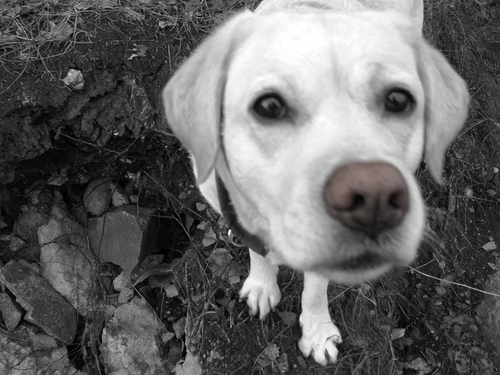}%
        }
        \resizebox{\linewidth}{!}{%
            \includegraphics[height=4cm]{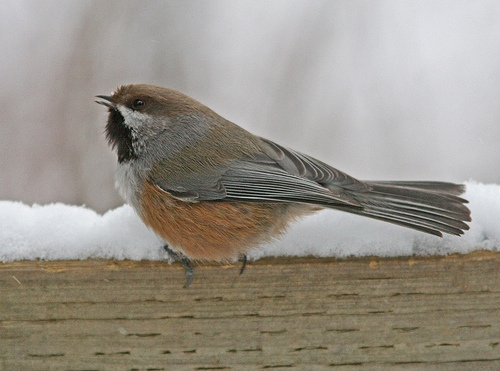}%
            \includegraphics[height=4cm]{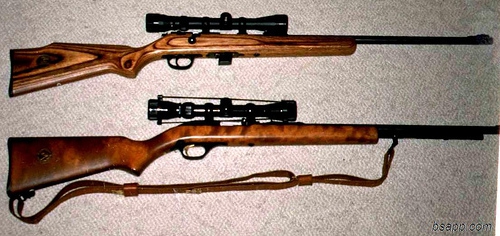}%
        }
        \caption{ImageNet}
    \end{subfigure}%
    \hfill%
    \begin{subfigure}[b]{.32\linewidth}%
        \offinterlineskip%
        \resizebox{\linewidth}{!}{%
            \includegraphics[height=4cm]{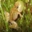}%
            \includegraphics[height=4cm]{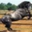}%
            \includegraphics[height=4cm]{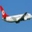}%
        }
        \resizebox{\linewidth}{!}{%
            \includegraphics[height=4cm]{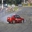}%
            \includegraphics[height=4cm]{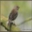}%
            \includegraphics[height=4cm]{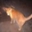}%
        }
        \caption{ciFAIR-10}
    \end{subfigure}%
    \hfill%
    \begin{subfigure}[b]{.288\linewidth}%
        \offinterlineskip%
        \resizebox{\linewidth}{!}{%
            \includegraphics[height=4cm]{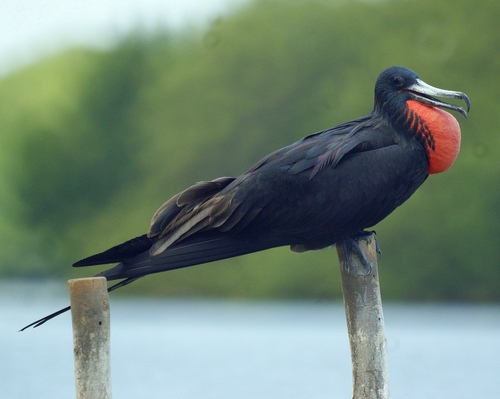}%
            \includegraphics[height=4cm]{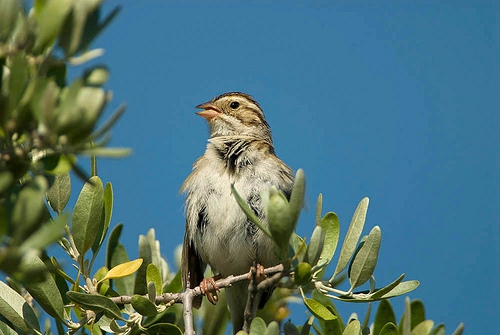}%
        }
        \resizebox{\linewidth}{!}{%
            \includegraphics[height=4cm]{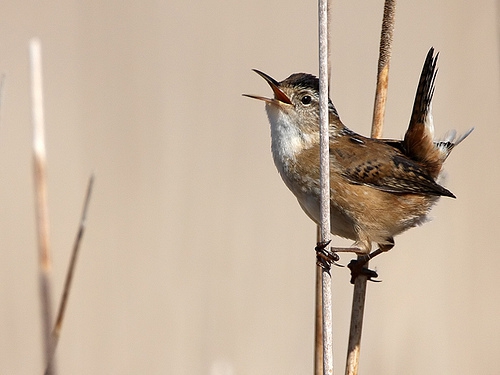}%
            \includegraphics[height=4cm]{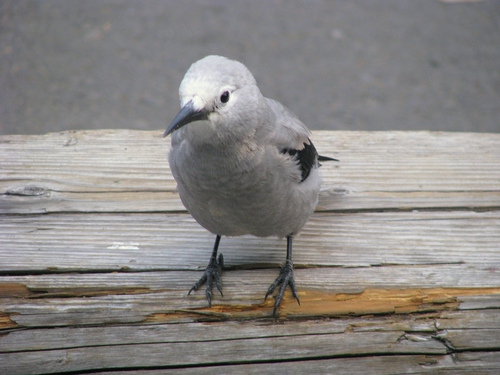}%
        }
        \caption{CUB}
    \end{subfigure}
    \\[.8\baselineskip]
    \begin{subfigure}[b]{.338\linewidth}%
        \offinterlineskip%
        \resizebox{\linewidth}{!}{%
            \includegraphics[height=4cm]{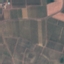}%
            \includegraphics[height=4cm]{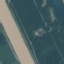}%
            \includegraphics[height=4cm]{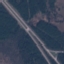}%
        }
        \resizebox{\linewidth}{!}{%
            \includegraphics[height=4cm]{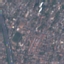}%
            \includegraphics[height=4cm]{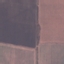}%
            \includegraphics[height=4cm]{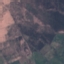}%
        }
        \caption{EuroSAT}
    \end{subfigure}%
    \hfill%
    \begin{subfigure}[b]{.3\linewidth}%
        \offinterlineskip%
        \resizebox{\linewidth}{!}{%
            \includegraphics[height=4cm]{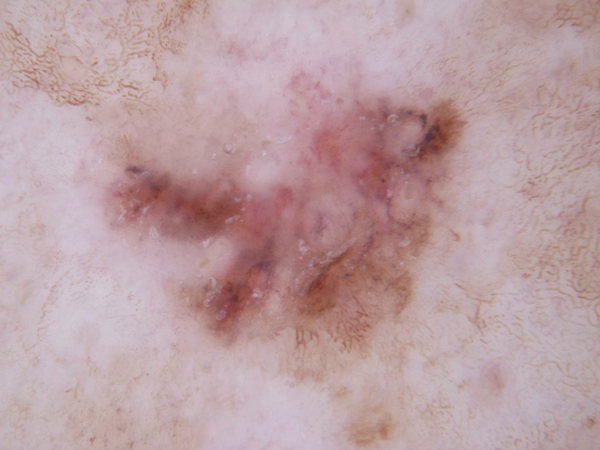}%
            \includegraphics[height=4cm]{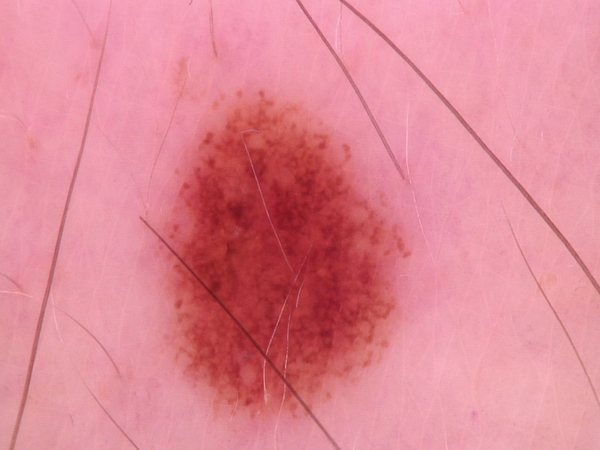}%
        }
        \resizebox{\linewidth}{!}{%
            \includegraphics[height=4cm]{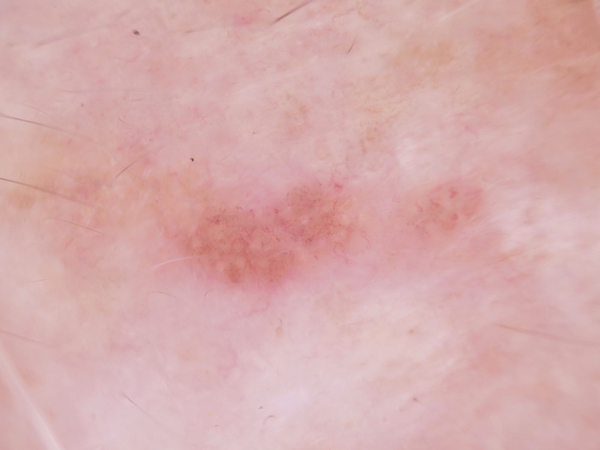}%
            \includegraphics[height=4cm]{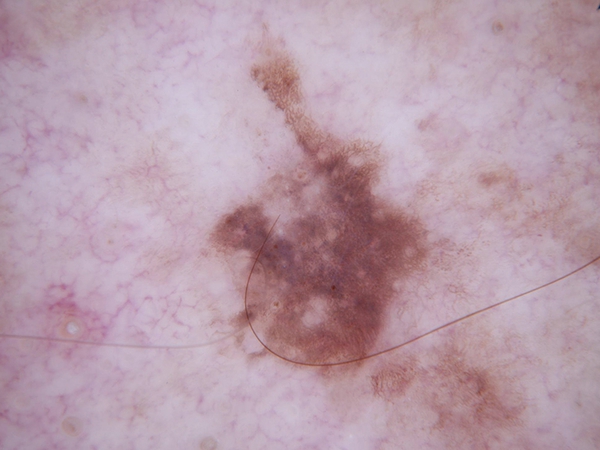}%
        }
        \caption{ISIC 2018}
    \end{subfigure}%
    \hfill%
    \begin{subfigure}[b]{.309\linewidth}%
        \offinterlineskip%
        \resizebox{\linewidth}{!}{%
            \includegraphics[height=4cm]{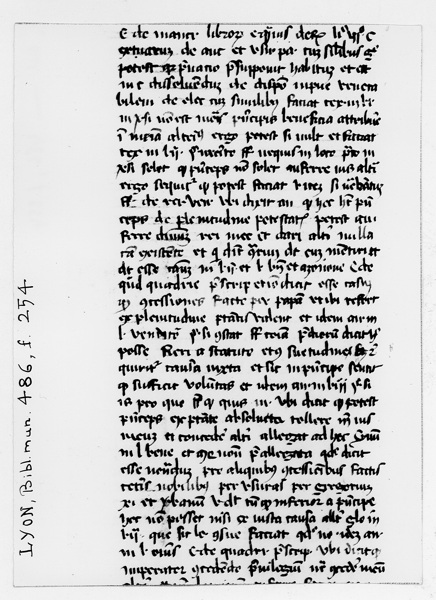}%
            \includegraphics[height=4cm]{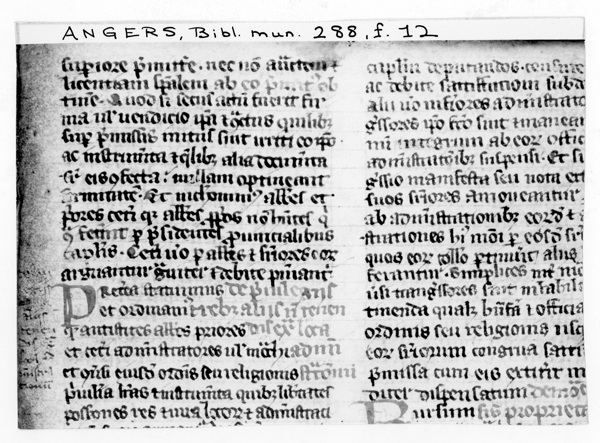}%
            \includegraphics[height=4cm]{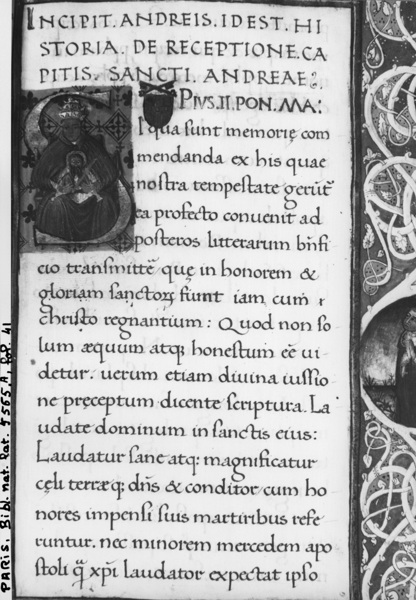}%
        }
        \resizebox{\linewidth}{!}{%
            \includegraphics[height=4cm]{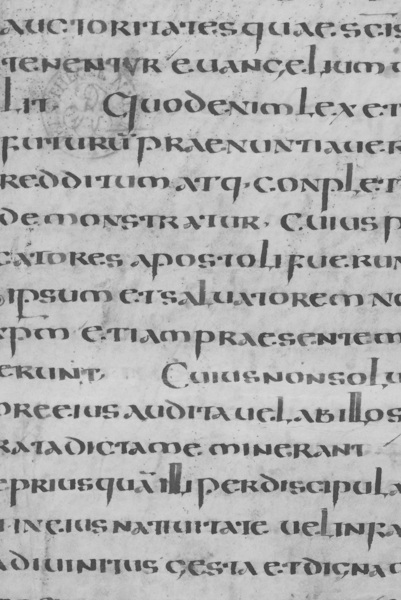}%
            \includegraphics[height=4cm]{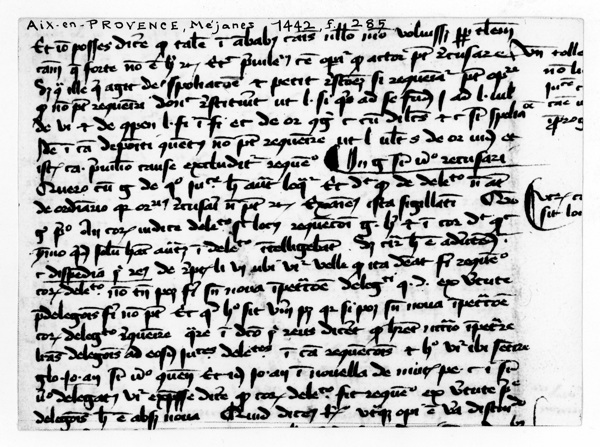}%
            \includegraphics[height=4cm]{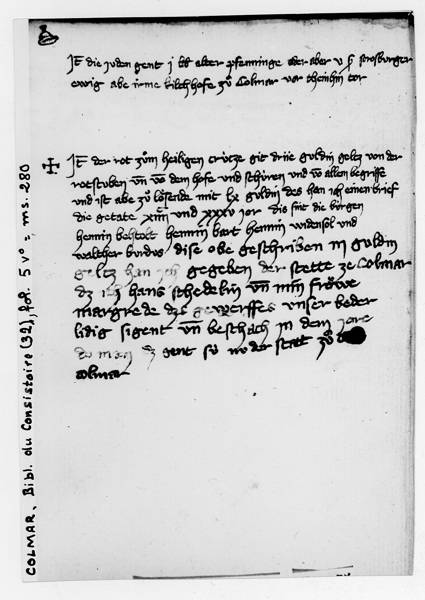}%
        }
        \caption{CLaMM}
    \end{subfigure}
    \caption{Example images from the datasets included in our benchmark. For EuroSAT, we only show the RGB bands.}
    \label{fig:dataset-examples}
\end{figure*}

\paragraph{ImageNet-1k \cite{russakovsky2015imagenet}} has been \emph{the} standard benchmark for image classification for almost a decade now and also served as a basis for challenge datasets for data-efficient image classification \cite{bruintjes2021vipriors}.
It comprises images from 1,000 classes of everyday objects and natural scenes collected from the web using image search engines.
Due to the large number of classes, a sub-sample of 50 images per class still results in a rather large training dataset compared with the rest of our benchmark.

\paragraph{ciFAIR-10 \cite{barz2020cifair}} is a variant of the popular CIFAR-10 dataset \cite{krizhevsky2009learning}, which comprises low-resolution images of size $32 \times 32$ from 10 different classes of everyday objects.
To a large part, its popularity stems from the fact that the low image resolution allows for fast training of neural networks and hence rapid experimentation.
However, the test dataset of CIFAR-10 contains about 3.3\% duplicates from the training set \cite{barz2020cifair}, which can potentially bias the evaluation.
The ciFAIR-10 dataset \cite{barz2020cifair} provides a variant of the test set, where these duplicates have been replaced with new images from the same domain.

\paragraph{Caltech-UCSD Birds-200-2011 (CUB) \cite{wah2011cub}} is a fine-grained dataset of 200 bird species.
Annotating this kind of images typically requires a domain expert and is hence costly.
Therefore, the dataset is rather small and only comprises 30 training images per class.
Pre-training on related large-scale datasets is hence the de-facto standard for CUB \cite{cui2018large,lin2015bilinear,Simon19:implicit,zheng2017learning}, which makes it particularly interesting for research on data-efficient methods closing the gap between training from scratch and pre-training.

\paragraph{EuroSAT \cite{helber2019eurosat}} is a multispectral image dataset based on Sentinel-2 satellite images of size 64x64 covering 13 spectral bands.
Each image is annotated with one of ten land cover classes.
This dataset does not only exhibit a substantial domain shift compared to standard pre-training datasets such as ImageNet but also a different number of input channels.
This scenario renders the standard pre-training and fine-tuning procedure impossible.

Nevertheless, Helber et al.~\cite{helber2019eurosat} adhere to this procedure by fine-tuning a CNN pre-trained on RGB images using different combinations of three out of the 13 channels of EuroSAT.
Unsurprisingly, they find that the combination of the R, G, and B channel provides the best performance in this setting.
This limitation to three channels due to pre-training is a waste of data and potential.
In our experiments on a smaller subset of EuroSAT, we found that using all 13 channels increases the classification accuracy by 9.5\% compared to the three RGB channels when training from scratch.

\paragraph{ISIC 2018 \cite{codella2019isic}} is a medical dataset consisting of dermoscopic skin lesion images, annotated with one of seven possible skin disease types.
Since medical data usually requires costly expert annotations, this domain is important to be covered by a benchmark on data-efficient deep learning.
Due to the small number of classes, we increase the number of images per class to 80 for this dataset, so that the size of the training set is more similar to our other datasets.

\paragraph{CLaMM \cite{stutzmann2016clamm}} is a dataset for \textbf{C}lassification of \textbf{La}tin \textbf{M}edieval \textbf{M}anuscripts.
It was originally used in the ICFHR 2016 Competition for Script Classification, where the task was to classify grayscale images of Latin scripts from handwritten books dated 500 C.E. to 1600 C.E. into one of twelve script style classes such as \emph{Humanistic Cursive}, \emph{Praegothica} etc.
This domain is quite different from that of typical pre-training datasets such as ImageNet and one can barely expect any useful knowledge to be extracted from ImageNet about medieval documents.
In addition, the standard pre-training and fine-tuning procedure would require the grayscale images to be converted to RGB for being passed through the pre-trained network, which incurs a waste of parameters.

\section{Methods}
\label{sec:methods}

In this section, we present the methods whose performance has been re-evaluated on our benchmark using the original code, where available.
We selected approaches for which the authors performed experiments on sub-sampled versions of standard computer vision datasets to prove their effectiveness for learning from small datasets.

\paragraph{Cross-entropy loss} is the widely used standard loss function for classification. We use it as a baseline with standard network architectures and optimization algorithms.

\paragraph{Deep hybrid networks (DHN)} represent one of the first attempts to incorporate pre-defined geometric priors via a hybrid approach of combining pre-defined and learned representations \cite{oyallon2017scaling, oyallon2018scattering}.
According to the authors, decreasing the number of parameters to learn could make deep networks more data-efficient, especially in settings where the scarcity of data would not allow the learning of low-level feature extractors.
Deep hybrid networks first perform a scattering transform on the input image generating feature maps and then apply standard convolutional blocks.  
The spatial scale of the scattering transform is controlled by the parameter \(J \in \mathbb{N}\).

\paragraph{Orthogonal low-rank embedding (OLÉ)} is a geometric loss for deep networks that was proposed in \cite{lezama2018ole} to reduce intra-class variance and enforce inter-class margins.
This method collapses deep features into a
learned linear subspace, or union of them, and inter-class subspaces are pushed to be as orthogonal as possible. The contribution of the low-rank embedding to the overall loss is weighted by the hyper-parameter \(\lambda_\mathrm{ole}\).

\paragraph{Grad-\(\mathbf{\ell_{2}}\) penalty} is a regularization strategy tested in the context of improving generalization on small datasets in \cite{bietti2019kernel}.
The \(\ell_{2}\) (squared) gradient norm is computed with respect to the input samples and used as a penalty in the loss  weighted by parameter \(\lambda_\mathrm{grad}\).
Among many regularization approaches evaluated in \cite{bietti2019kernel}, we have chosen the grad-\(\ell_{2}\) penalty because it was among the best performing methods in the experiments with ResNet and sub-sampled versions of CIFAR-10.
Since the grad-\(\ell_{2}\) penalty is proposed as an alternative to weight decay, we disable weight decay for this method.
Moreover, differently from the original implementation, we enabled the use of batch normalization since, without this component, we obtained extremely low results in preliminary experiments.

\paragraph{Cosine loss} was proposed in \cite{barz2020deep} to decrease overfitting in problems with scarce data.
Thanks to an $\ell_2$ normalization of the learned feature space, the cosine loss is invariant against scaling of the network output and solely focuses on the directions of feature vectors instead of their magnitude.
In contrast to the softmax function used with the cross-entropy loss, the cosine loss does not push the activations of the true class towards infinity, which is commonly considered as a cause of overfitting \cite{szegedy2016rethinking,he2018bag}.
Moreover, a further increase of performance was obtained by combining the cosine with the cross-entropy loss after an additional layer on top of the embeddings learned with the cosine loss.

\paragraph{Harmonic networks (HN)} use a set of preset filters based on windowed cosine transform at several frequencies which are combined by learnable weights \cite{ulicny2019harmonicnet, ulicny2019harmonic}.
Similar to hybrid networks, the idea of the harmonic block is to have a useful geometric prior that can help to avoid overfitting.
Harmonic networks use Discrete Cosine Transform filters which have excellent energy compaction properties and are widely used for image compression.

\paragraph{Full convolution (F-Conv)} was proposed in \cite{kayhan2020translation} to improve translation invariance of convolutional filters.
Standard CNNs exploit image boundary effects and learn filters that can exploit the absolute spatial locations of objects in images.
In contrast, full convolution applies each value in the filter on all values in the image.
According to \cite{kayhan2020translation}, improving translation invariance strengthens the visual inductive prior of convolution, leading to increased data efficiency in the small-data setting.

\paragraph{Dual Selective kernel networks} have been proposed and designed in \cite{sun2020visual} to be more data-efficient.
The standard residual block is modified, keeping the skip connection, with two forward branches that use \(\ 1 \times 1\) convolutions, selective kernels \cite{li2019selective} and an anti-aliasing module.
To further regularize training, only one of the two branches is randomly selected in the forward and backward passes, while at inference, the two paths are weighted equally.

Besides the specialized network architecture, the original work uses a combination of three custom loss functions \cite{sun2020visual}.
Despite best efforts, we were unable to derive the correct implementation from the ambiguous description of these loss functions in the paper.
Therefore, we only use the DSK network architecture with cross-entropy loss.

\paragraph{T-vMF Similarity} is a generalization of the cosine similarity that was recently presented in \cite{kobayashi2021t} to make modern CNNs more robust to some realistic learning situations such as class imbalance, few training samples, and noisy labels.
As the name suggests, this similarity is mainly based on the von Mises-Fisher distribution of directional statistics and built on top of the heavy-tailed student-t distribution.
The combination of these two ingredients provides high compactness in high-similarity regions and low similarity in heavy-tailed ones.
The degree of compactness/dispersion of the similarity is controlled by the parameter \(\kappa\).

\section{Experimental setup}
\label{sec:setup}

In this section, we give an overview of the experimental pipeline that we followed for a fair evaluation of the aforementioned methods on the six datasets that constitute our benchmark.

\subsection{Evaluation metrics}
In our benchmark, we evaluate each method on each dataset with the widely used balanced classification accuracy.
This metric is defined as the average per-class accuracy, \ie, the average of the diagonal in the row-normalized confusion matrix.
We turned our attention toward this metric since some datasets in our benchmark do not have balanced test sets.
In any case, for balanced test sets, the balanced accuracy equals the standard classification accuracy.

Since our benchmark contains multiple datasets it is hard to directly make a comparison between two methods without computing an overall ranking.
Therefore, for each method, we also compute the average balanced accuracy across all datasets to provide a simple and intuitive way to rank methods.
Additionally, in this manner, future methods will be easily comparable with those already evaluated.

\subsection{Data pre-processing and augmentation} 
All input images were normalized by subtracting the channel-wise mean and dividing by the standard deviation computed on the \textit{trainval} split.
We applied standard data augmentation policies with slightly varying configurations, adapted to the specific characteristics of each dataset and problem domain.
Note that none of the currently re-evaluated methods in our benchmark had as original contribution a specialized data augmentation technique.
Nothing prevents the use of an augmentation-based method from partaking in the benchmark.

For datasets with a small, fixed image resolution, \ie, ciFAIR-10 and EuroSAT, we perform random shifting by 12.5\% of the image size and horizontal flipping in 50\% of the cases.
For all other datasets, we apply scale augmentation using the \texttt{RandomResizedCrop} transform from PyTorch\footnote{\url{https://pytorch.org/vision/stable/transforms.html\#torchvision.transforms.RandomResizedCrop}} as follows:
A crop with a random aspect ratio drawn from $[\frac{3}{4}, \frac{4}{3}]$ and an area between $A_\mathrm{min}$ and 100\% of the original image area is extracted from the image and then resized to $224 \times 224$ pixels.
The minimum fraction $A_\mathrm{min}$ of the area was determined based on preliminary experiments to ensure that a sufficient part of the image remains visible.
It therefore varies depending on the dataset: We use $A_\mathrm{min} = 10\%$ for ImageNet, $A_\mathrm{min} = 20\%$ for CLaMM and $A_\mathrm{min} = 40\%$ for CUB and ISIC 2018.

For ISIC 2018 and EuroSAT, we furthermore perform random vertical flipping in addition to horizontal flipping, since these datasets are completely rotation-invariant and vertical reflection augments the training sets without drifting them away from the test distributions.
On CLaMM, in contrast, we do not perform any flipping, since handwritten scripts are not invariant even against horizontal flipping.

\begin{table*}[t]
    \renewcommand{\arraystretch}{1.3}
    \setlength{\tabcolsep}{10pt}
    \begin{tabularx}{\linewidth}{X cccccc }
        \toprule
        Hyper-Parameter &    ImageNet    &   ciFAIR-10    &      CUB       &    EuroSAT     &   ISIC 2018    &     CLaMM\\
        \midrule
        Learning Rate   & \multicolumn{6}{c}{\texttt{loguniform}(1e-4, 0.1)}  \\
        \rowcolor{LightGray}
        Weight Decay    & \multicolumn{6}{c}{\texttt{loguniform}(1e-5, 0.1)} \\
        Batch Size      &         \{8, 16, 32\}  &  \{10, 25, 50\}  &   \{8, 16, 32\}  &  \{10, 25, 50\}  &   \{8, 16, 32\}   &   \{8, 16, 32\}\\
        \rowcolor{LightGray}
        Epochs          &   200 (500)  &    500  &   200  & 500  & 500  & 500 \\
        HPO Trials      & 100  &   250  &   100  &   250  &  100  & 100\\
        \rowcolor{LightGray}
        Grace Period    &     10  &   50  &   10  &  25  &  25  & 25\\
        \bottomrule
    \end{tabularx}
    \caption{Summary of hyper-parameters searched/used with ASHA \cite{li2020system}. Method specific hyper-parameters were included in the search space but not included in this table due to space limitations. An epoch number in parentheses means that a higher number of epochs was used for the final training than for the hyper-parameter optimization.}
    \label{tab:hpo_params}
\end{table*}

\subsection{Architecture and optimizer}
To perform a fair comparison, we use the same backbone CNN architecture for all methods.
More precisely, for ciFAIR-10, we employ a Wide Residual Network (WRN) \cite{zagoruyko2016wrn}, precisely WRN-16-8, which is widely used in the existing literature for data-efficient classification on CIFAR.
For all other cases, the popular and well-established ResNet-50 (RN50) architecture \cite{he2016deep} is used.
Note that we made changes to the architecture when that was an original contribution of the paper, but all those changes were applied to the selected base architecture.
Due to the high popularity of residual networks, the majority of the selected approaches were originally tested with a RN/WRN backbone.
This fact allowed us to perform a straightforward porting of the network setup, when necessary.

We furthermore employ a common optimizer and training schedule across all methods and datasets to avoid any kind of optimization bias.
We use standard stochastic gradient descent (SGD) with a momentum of 0.9 and weight decay and a cosine annealing learning rate schedule \cite{loshchilov2016sgdr}, which reduces the learning rate smoothly during the training process.
The initial learning rate and the weight decay factor are optimized for each method and dataset individually together with any method-specific hyper-parameters as detailed in the next subsection.
The total number of training epochs for each dataset was chosen according to preliminary experiments.

\subsection{Hyper-parameter optimization}
As we have discussed in \cref{sec:intro} and shown in \cref{fig:gridsearch}, the choice of hyper-parameters has a substantial effect on the classification performance that should in no case be underestimated.
Careful hyper-parameter optimization (HPO) \cite{bischl2021hpo} is therefore not only crucial for applying deep learning techniques in practice but also for a fair comparison between different methods, so that each can obtain its optimal performance.
Comparing against an untuned baseline with default hyper-parameters is as good as no comparison at all.

For our benchmark, we hence first tune the hyper-parameters of each method on each individual dataset using a training and a validation split, which are disjoint from the test set used for final performance evaluation (see \cref{sec:datasets}).
For any method, we tune the initial learning rate and weight decay, sampled from a log-uniform space, as well as the batch size, chosen from a pre-defined set.
Details about the search space are provided in \cref{tab:hpo_params}.
In addition to these general hyper-parameters, any method-specific hyper-parameters are tuned as well simultaneously, considering the boundaries used in the original paper, if applicable, or lower and upper bounds estimated by ourselves.

For selecting hyper-parameters to be tested and scheduling experiments, we employ Asynchronous HyperBand with Successive Halving (ASHA) \cite{li2020system} as implemented in the Ray library\footnote{\url{https://docs.ray.io/en/master/tune/}}.
This search algorithm exploits parallelism and aggressive early-stopping to tackle large-scale hyper-parameter optimization problems.
Trials are evaluated and stopped based on their accuracy on the validation split.

Two main parameters need to be configured for the ASHA algorithm: the number of trials and the grace period.
The former controls the number of hyper-parameter configurations tried in total while the latter the minimum time after which a trial can be stopped.
Since the number of trials corresponds to the time budget available for HPO, we choose larger values for smaller datasets, where training is faster.
The grace period, on the other hand, should be large enough to allow for a sufficient number of training iterations before comparing trials.
Therefore, we choose larger grace periods for smaller datasets, where a single epoch comprises fewer training iterations.
The exact values for each dataset as well as the total number of training epochs can be found in \cref{tab:hpo_params}.
These values were determined based on preliminary experiments with the cross-entropy baseline.

\subsection{Final training and evaluation}
After having completed HPO using the procedure described above, we train the classifier with the determined configuration on the combined training and validation split and evaluate the balanced classification accuracy on the test split.
To account for the effect of random initialization, this training is repeated ten times and we report the balanced average accuracy.

\begin{table*}[t]
    \renewcommand{\arraystretch}{1.3}
    \setlength{\tabcolsep}{5pt}
    \begin{tabularx}{\linewidth}{X cccccc c}
        \toprule
        Method                                                               &    ImageNet    &   ciFAIR-10    &      CUB       &    EuroSAT     &   ISIC 2018    &     CLaMM      &   Average   \\
        \midrule
        Cross-Entropy Baseline                                               &         44.97  & \textbf{58.22} &         71.44  &         90.27  &         67.19  & \textbf{75.34} &          67.90  \\
        \rowcolor{LightGray}
        Deep Hybrid Networks \cite{oyallon2017scaling,oyallon2018scattering} &         38.69  &         54.21  &         52.54  &         91.15  &         59.64  &         65.74  &          60.33  \\
        OL{\'E} \cite{lezama2018ole}                                         &         43.05  &         54.92  &         63.32  &         89.29  &         62.89  &         71.42  &          64.15  \\
        \rowcolor{LightGray}
        Grad-\(\ell_{2}\) Penalty \cite{bietti2019kernel}                       &         25.21  &         51.03  &         51.94  &         79.33  &         60.21  &         65.10  &          55.47  \\
        Cosine Loss \cite{barz2020deep}                                      &         37.22  &         52.39  &         66.94  &         88.53  &         62.42  &         68.89  &          62.73  \\
        \rowcolor{LightGray}
        Cosine Loss + Cross-Entropy \cite{barz2020deep}                      &         44.39  &         51.74  &         70.80  &         88.77  &         64.52  &         69.29  &          64.92  \\
        Harmonic Networks \cite{ulicny2019harmonicnet,ulicny2019harmonic}    & \textbf{46.36} &         56.50  & \textbf{72.26} & \textbf{92.09} & \textbf{70.42} & \textit{74.59} &  \textbf{68.70} \\
        \rowcolor{LightGray}
        Full Convolution \cite{kayhan2020translation}                        &         36.58  &         55.00  &         64.90  &         90.82  &         61.70  &         63.33  &          62.06  \\
        Dual Selective Kernel Networks \cite{sun2020visual}                  &         45.21  &         54.06  &         71.02  &         91.25  &         64.78  &         61.51  &          64.64  \\
        \rowcolor{LightGray}
        T-vMF Similarity \cite{kobayashi2021t}                               &         42.79  & \textit{57.50} &         67.43  &         88.53  &         65.37  &         66.40  &          64.67  \\
        \bottomrule
    \end{tabularx}
                    
    \caption{Average balanced classification accuracy in \% over 10 runs for each task and across all tasks. The best value per dataset is highlighted in bold font. Numbers in italic font indicate that the result is not significantly worse than the best one on a significance level of 5\%.}
    \label{tab:results}
\end{table*}

\section{Results}
\label{sec:results}

In the following, we first present the results of re-evaluating the eight methods described in \cref{sec:methods} and the baseline on our benchmark introduced in \cref{sec:datasets}, after carefully tuning all methods on each dataset.
Then, we compare the performance obtained by our re-implementations, including the baseline, with other values published in the literature.

\subsection{Data-efficient image classification benchmark}

\Cref{tab:results} presents the average balanced classification accuracy over 10 runs with different random initializations for all methods and datasets.
We performed Welch's t-test to assess the significance of the advantage of the best method for each dataset in comparison to all others.
Most results are significantly worse on a level of 5\% than the best method on the respective dataset, with only two exceptions: T-vMF Similarity on ciFAIR-10 and Harmonic Networks on CLaMM perform similar to the best method on these datasets, which is the baseline in both cases.

This leads us to the main surprising finding of this benchmark:
When tuned carefully, the standard cross-entropy baseline is very competitive with the published methods specialized for deep learning from small datasets.
On ciFAIR-10 and CLaMM, it actually is the best performing method.
It obtains the second rank on CUB and ISIC 2018, the third rank on ImageNet, and the fourth on EuroSAT.
The baseline scores an average balanced accuracy across all datasets of 67.90\%, which beats all other methods except Harmonic Networks by a large margin (the next best average accuracy is only 64.92\%).

Harmonic Networks are the overall champion of our benchmark, with an average balanced accuracy of 68.70\%.
On the four datasets where they outperform the baseline, however, they only surpass it by 1\%-5\%.

Overall, the finding that the vast majority of recent methods for data-efficient image classification does not even achieve the same performance as the baseline is sobering.
We attribute this to the fact that the importance of hyper-parameter optimization is immensely underestimated, resulting in misleading comparisons of novel approaches with weak and underperforming baselines.

\begin{table*}[h]
    \renewcommand{\arraystretch}{1.3}
    \setlength{\tabcolsep}{5pt}
    \begin{tabularx}{\textwidth}{ccccXcccc}
        \toprule
        \multicolumn{4}{c}{Cross-Entropy Baseline} &  & \multicolumn{4}{c}{Other Methods} \\ 
        \cmidrule{1-4} \cmidrule{6-9}
        Publication  & Dataset   &  Network & Accuracy  &  & Method & Dataset & Network & Accuracy\\
        \cmidrule{1-4} \cmidrule{6-9}

         \cite{oyallon2017scaling}  & CIFAR-10  & WRN-16-8  &  46.5 $\pm$ 1.4 & & {DHN \cite{oyallon2017scaling}}\cellcolor{LightGray} & CIFAR-10 \cellcolor{LightGray} & WRN-16-8 \cellcolor{LightGray} & 54.7 $\pm$ 0.6 \cellcolor{LightGray}\\
         
         \cite{ulicny2019harmonic} & CIFAR-10 &  WRN-16-8  & 52.2 $\pm$ 1.8 & & DHN (Ours)  \cellcolor{LightGray}  & ciFAIR-10 \cellcolor{LightGray}& WRN-16-8 \cellcolor{LightGray}&  54.21 $\pm$ 0.4 \cellcolor{LightGray}\\
         
         Ours  & ciFAIR-10  &  WRN-16-8  &  58.22 $\pm$ 0.9  & & HN \cite{ulicny2019harmonic}  & CIFAR-10   & WRN-16-8   & 58.4 $\pm$ 0.9 \\
         
        \cite{kayhan2020translation} \cellcolor{LightGray} & ImageNet \cellcolor{LightGray} & RN50 \cellcolor{LightGray} &  26.39 \cellcolor{LightGray} & & HN (Ours)   & ciFAIR-10 & WRN-16-8 &  56.50 $\pm$ 0.5 \\
        
        Ours \cellcolor{LightGray}& ImageNet \cellcolor{LightGray}& RN50 \cellcolor{LightGray} &  44.97 $\pm$ 0.3 \cellcolor{LightGray} & & F-Conv \cite{kayhan2020translation} \cellcolor{LightGray}  & ImageNet \cellcolor{LightGray}& RN50 \cellcolor{LightGray} &  31.1 \cellcolor{LightGray}\\
        
         &  &  &   & & F-Conv (Ours) \cellcolor{LightGray}   & ImageNet \cellcolor{LightGray}& RN50 \cellcolor{LightGray}&  36.58 $\pm$ 0.4 \cellcolor{LightGray}\\

        \bottomrule
    \end{tabularx}
    \caption{Summary of ours/published results of the cross-entropy baseline (left) and other methods (right) on similar setups. }
    \label{tab:baseline-comparison}
\end{table*}

\subsection{Published baselines are underperforming}

We show further evidence of why tuning the hyper-parameters and not neglecting the baseline in scenarios with small datasets is fundamental to perform a fair comparison between different methods.

We analyzed the original results reported for the methods considered in our study and selected those that shared a similar setup.
Note that due to the lack of a standard benchmark and the common practice of randomly sub-sampling large datasets, we are unable to conduct a fair comparison with the same dataset split, training procedure, etc.
Still, our benchmark shares the base dataset and network architecture with the selected cases.
Therefore, we believe that this analysis is suitable for supporting our point regarding the common practice of comparing tuned proposed methods with underperforming baselines.

The results of this analysis are shown in \cref{tab:baseline-comparison}.
Deep Hybrid Networks and Harmonic Networks were originally tested with a WRN-16-8 on CIFAR-10 while Full Convolution employed RN50 on ImageNet.
In both cases, training sets were comprised of \(50\) images per class.
Our baseline clearly outperforms the original baselines by large margins (\cref{tab:baseline-comparison}, left part).
More precisely, our models surpass the reported ones by \textapprox 12, \textapprox 6, and \textapprox 18 percentage points on the CIFAR and ImageNet setups.
Recall also that the ciFAIR-10 test set is slightly harder than the CIFAR-10 one due to the removal of duplicates \cite{barz2020cifair}.

On the contrary, for the case of the proposed methods (\cref{tab:baseline-comparison}, right part), the difference between ours and original results is sharply less evident.
Our DHN and HN slightly underperform the original ones by a \textapprox 0.5 and \textapprox 2 percentage points, respectively.
However, this was expected due to the higher difficulty of ciFAIR-10. 
On ImageNet instead, our F-Conv model outperforms the original one by \textapprox 5 percentage points, confirming once again that careful HPO can further boost the performance.

From this analysis it seems clear that proposed methods are usually tuned to obtain an optimal or near-optimal result while baselines are trained with default hyper-parameters that have been found useful for large datasets but do not necessarily generalize to smaller ones.

\subsection{Optimal hyper-parameters}


\begin{figure}[t]
    \includegraphics[width=\linewidth]{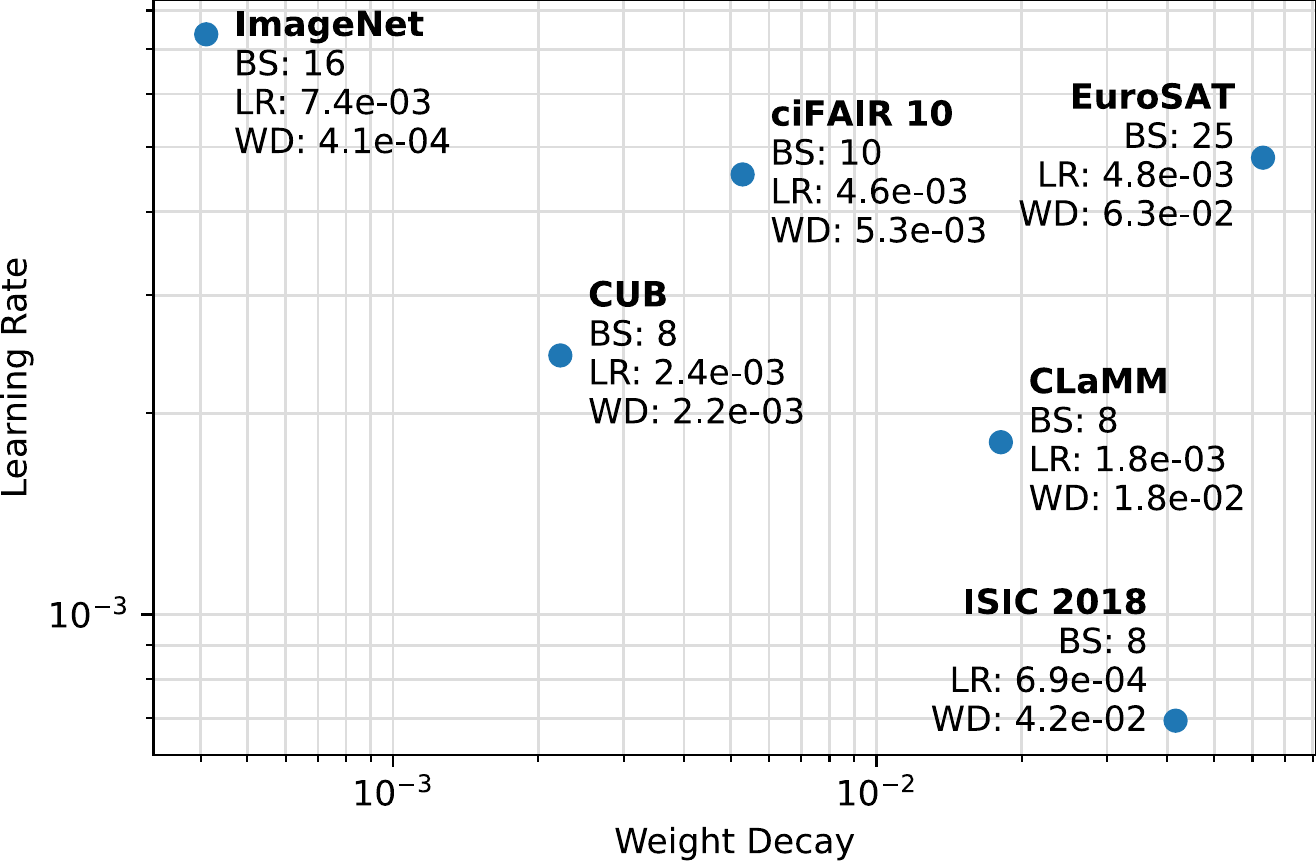}
    \caption{Hyper-parameters found with ASHA \cite{li2020system} for the cross-entropy baseline. BS = batch size, LR = learning rate, WD = weight decay.}
    \label{fig:hparams}
\end{figure}

For reproducibility, but also to gain further insights into hyper-parameter optimization for small datasets, we show the best hyper-parameter combinations found during our search for the cross-entropy baseline in \cref{fig:hparams}.

We can observe that small batch sizes seem to be beneficial, despite the use of batch normalization.
While the learning rate exhibits a rather small range of values from \num{0.7e-3} to \num{7.4e-3} across datasets and spans only one order of magnitude, weight decay varies within a range of two orders of magnitude from \num{4.1e-4} to \num{1.8e-2}.

Furthermore, learning rate and weight decay appear to be negatively correlated.
Higher learning rates are usually accompanied by smaller weight decay factors.
The same correlation can be observed in \cref{fig:gridsearch}.
A quantitative analysis over the hyper-parameters of all methods used in our study instead of only the baseline yields a correlation of $r = -.28,\ p = .02$.
After taking the logarithm of learning rate and weight decay, the correlation is strengthened to $r = -.58,\ p < .01$.

\section{Conclusions}
\label{sec:conclusions}

In this paper, we laid the foundation for fair and appropriate comparisons among modern data-efficient image classifiers.
The motivations that brought us to our work are mainly two-fold: the lack of a common evaluation benchmark with fixed datasets, architectures, and training pipelines; and the experimental evidence of weak assessments of baselines due to a lack of careful tuning.

The re-evaluation of eight selected state-of-the-art methods guided us to the surprising and sobering conclusion that the standard cross-entropy loss ranks second in our benchmark only behind Harmonic Networks and, competes with or outperforms the remaining methods.

With these results in mind, we conclude that the importance of hyper-parameter  optimization  is  immensely  undervalued and should be taken into account in future studies to elude misleading comparisons of new approaches with weak and underperforming baselines.
The publication of our benchmark heads towards this direction and is considered by ourselves an important contribution for the community of data-efficient image classification.

{\small
\bibliographystyle{ieee_fullname}
\bibliography{egbib}
}

\end{document}